\documentclass[sigconf,authorversion,nonacm]{acmart}

\makeatletter
\def\@ACM@checkaffil{
    \if@ACM@instpresent\else
    \ClassWarningNoLine{\@classname}{No institution present for an affiliation}%
    \fi
    \if@ACM@citypresent\else
    \ClassWarningNoLine{\@classname}{No city present for an affiliation}%
    \fi
    \if@ACM@countrypresent\else
        \ClassWarningNoLine{\@classname}{No country present for an affiliation}%
    \fi
}
\makeatother

\usepackage{balance}
\usepackage{soul}
\usepackage{url}
\usepackage[utf8]{inputenc}
\usepackage{graphicx}
\usepackage{amsmath}
\usepackage{amsthm}
\usepackage{booktabs}
\usepackage{algorithm}
\usepackage[noend]{algorithmic}
\usepackage{comment}
\usepackage{color}
\usepackage{appendix}
\usepackage{multirow}

\newcommand{\citeit}[1]{\citeauthor{#1}~(\citeyear{#1})}

\newtheorem{definition}{Definition}
\newtheorem{theorem}{Theorem}

\newtheorem{lemma}{Lemma}[theorem]

\title{Causal Explanations for Sequential Decision Making\\ Under Uncertainty}

\author{Samer B. Nashed}
\affiliation{
  \institution{University of Massachusetts Amherst}}
\email{snashed@cs.umass.edu}

\author{Saaduddin Mahmud}
\affiliation{
  \institution{University of Massachusetts Amherst}}
\email{smahmud@cs.umass.edu}

\author{Claudia V. Goldman}
\affiliation{
  \institution{General Motors Research}}
\email{claudia.goldman@gm.com}

\author{Shlomo Zilberstein}
\affiliation{
  \institution{University of Massachusetts Amherst}}
\email{shlomo@cs.umass.edu}

\begin{document}

\begin{abstract}
We introduce a novel framework for causal explanations of stochastic, sequential decision-making systems built on the well-studied structural causal model paradigm for causal reasoning. This single framework can identify multiple, semantically distinct explanations for agent actions --- something not previously possible. In this paper, we establish exact methods and several approximation techniques for causal inference on Markov decision processes using this framework, followed by results on the applicability of the exact methods and some run time bounds. We discuss several scenarios that illustrate the framework's flexibility and the results of experiments with human subjects that confirm the benefits of this approach.
\end{abstract}

\maketitle

\section{Introduction}

As autonomous decision making becomes ubiquitous, researchers agree that developing trust is required for adoption and proficient use of AI systems~\cite{linegang2006human,stubbs2007autonomy,zhang2020effect}, and it is widely accepted that autonomous agents that can explain their decisions help promote trust~\cite{chen2018situation,hayes2017improving,mercado2016intelligent}. However, there are many challenges in generating such explanations. Consider, for example, an autonomous vehicle (AV) stopping behind a truck for a long duration. The passenger may wonder whether the AV is waiting for the truck to move, waiting for an opportunity to pass the truck, or dealing with some technical problem. Generating suitable explanations of such a system is hard due to the complexity of planning, which may involve large state spaces, stochastic actions, imperfect observations, and complicated objectives. Furthermore, useful explanations must somehow reduce the internal reasoning process to a form understandable by a user who likely does not know all of the algorithmic details. 

Another challenging aspect is the heterogeneity of possible operational contexts and interaction with different types of users with different expectations. For example, in the above AV scenario, the explanation furnished to a passenger may differ from that given a driver evaluating whether or not to intervene and take control of the vehicle. Moreover, different planning, learning, and decision-making algorithms may not afford universal mechanisms for explanation due to fundamental differences in available information. 

Debate on the definition, taxonomy, and purpose of explanations has been well-represented in the cognitive science, psychology, and philosophy literature for a long time. While still active, there are several insights for which there is broad consensus \cite{miller2019explanation,mittelstadt2019explaining}, and we use this knowledge to motivate our approach. Scholars studying explanations mostly agree that requests for explanations are often motivated by a mismatch between the mental model of the requester and a logical conclusion based on observation~\cite{heider1958psychology,hesslow1988problem,hilton1986knowledge,hilton1996mental,lombrozo2006structure,williams2013hazards}, which creates a form of generalized model reconciliation problem~\cite{chakraborti2017plan}. Researchers also agree that explanations often require counterfactual analysis \cite{mackie1980cement,lipton1990contrastive,hilton1990conversational,lombrozo2012explanation}, which in turn requires causal determination \cite{woodward2005making,salmon2006four,lombrozo2010causal}. There are several computational paradigms for causal analysis, including those based on conditional logic \cite{lewis1974causation,giordano2004conditional}, and statistics \cite{freedman2007statistical}. Among the most well-studied paradigms is the \emph{structural causal model} (SCM) \cite{halpern2000causes}.

The primary contribution of this paper is a framework, based on SCMs, for applying causal analysis to sequential decision-making agents. We create an SCM representing the computation needed to derive a policy for a Markov decision process (MDP) and apply causal inference to identify variables that cause certain agent behavior. These variables can then be used to generate explanations, for example by completing natural language templates. This framework provides two main benefits. First, it is theoretically sound, based on concepts and formalisms from the causality literature, while most existing approaches use heuristics. Second, it is more flexible, allowing us to identify multiple types of explanans, whereas existing approaches often explain events in terms of a single set of variables in the decision-making model. For example, they may use \emph{only} state factors or \emph{only} reward variables, whereas we may use any set, increasing our framework's applicability. Furthermore, we offer several approximate techniques for large problems or problems where the topology of the causal graph prevents exact inference.

Theoretically, we determine the domain of problems for which this method is exact and provide some worst-case run time bounds for the algorithms presented. Empirically, we discuss several qualitative scenarios that illustrate how our approach not only produces sensible causes of agent behavior, but also uses a single framework to identify causal variables from a variety of MDP components. This case study is presented along with results from a user study in which users compare the proposed method to existing, heuristic methods and we find statistically significant preferences in favor of explanations generated via causal reasoning.

\section{Related Work}

Automatically generating explanations is a growing field of AI research. One focus area aims to explain the decisions of black-box machine learning algorithms \cite{mothilal2020explaining,lucic2020does,karimi2021algorithmic}. These works often use the terms explainable or interpretable machine learning (XML). Another focus area, more aligned with this work, aims to explain the outputs of planning algorithms, or modify planning algorithms so that they produce plans that are inherently more explainable. These works typically use the term explainable planning (XAIP). A large portion of XAIP research has been devoted to deterministic planners, or analyzing plans after they have been executed \cite{fox2017explainable,chakraborti2019explicability,chakraborti2020emerging}. However, many applications operate in stochastic domains or require explanations in real time.

Research on explanations of stochastic planners, such as MDPs, is relatively sparse, but there are several notable existing efforts. \citeit{elizalde2009generating} identify important state factors by looking at how the value function would change were they to perturb that state factor's value, and Khan et al.~(\citeyear{khan2009minimal}) present a technique to explain policies for factored MDPs by analyzing the expected occupancy frequency of states with extreme reward values. Similarly, Juozapaitis et al.~(\citeyear{juozapaitis2019explainable}) analyze how extreme reward values impact action selection in decomposed-reward RL agents, and Bertram and Peng~(\citeyear{bertram2018explainable}) look at reward sources in deterministic MDPs. Wang et al.~(\citeyear{wang2016impact}) try to explain policies of partially observable MDPs by communicating the relative likelihoods of different events or levels of belief. However, research clearly indicates that humans are not good at using this kind of numerical information~\cite{miller2019explanation}.

While these approaches have limited scope in the explanations they provide, they are also computationally cheap and easy to implement. More complex methods have been developed that attempt explanation via summarization. Pouget et al.~(\citeyear{pouget2020ranking}) identify key state-action pairs via spectrum-based fault localization, and Russell et al.~(\citeyear{russell2019explaining}) use decision trees to approximate a given policy and analyze the decision nodes to determine which state factors are most influential for immediate reward. Panigutti et al.~(\citeyear{panigutti2020doctor}) used similar methods to explain classifiers.

Recently, the use of structural causal models for explaining MDPs has been proposed by Madumal et al.~(\citeyear{madumal2020explainable}), who used SCMs to encode the influence of particular actions available to the agent. This approach was used in a model-free, reinforcement learning setting to learn the structural equations as multivariate regression models during training. However, it requires several strong assumptions including the prior availability of a graph representing causal direction between variables, discrete actions, and the existence of sink states. In contrast, our proposed framework allows causal analysis of all the components of MDPs using a single set of algorithms. Moreover, it is theoretically well-justified as it rests on a concrete theory of causality and can be easily extended for cases where approximate reasoning is required, including model-free planners.

\vspace{-1mm}
\section{Background}

Here, we provide an overview of concepts and notation for structural causal models and Markov decision processes.

\vspace{-1mm}
\subsection{Structural Causal Models}

Structural causal models (SCMs) \cite{halpern2000causes,halpern2005causes} are defined as tuple $\mathcal{S} = \langle \mathcal{U}, \mathcal{V}, \mathcal{M} \rangle$, describing a causal reasoning problem. The set $\mathcal{U}$, called the \emph{context}, is a set of exogenous variables that describe some condition of the world. These variables, while possibly causally relevant, are assumed to be fixed for a given analysis. The set $\mathcal{V}$ is a set of endogenous variables considered possible causes of some event. In our case, these variables are internal to the reasoning process of the agent. All variables in the world are either elements of $\mathcal{U}$ or elements of $\mathcal{V}$, and $\mathcal{U} \cap \mathcal{V} = \emptyset$. The final component, $\mathcal{M}$, is a set of equations that specify the causal effects of variables in $\mathcal{U}$ and $\mathcal{V}$ on other variables in $\mathcal{V}$. The decision of which variables to put into each set is a design choice that we revisit later. 

\begin{figure}[t]
    \centering
    \includegraphics[width=3in]{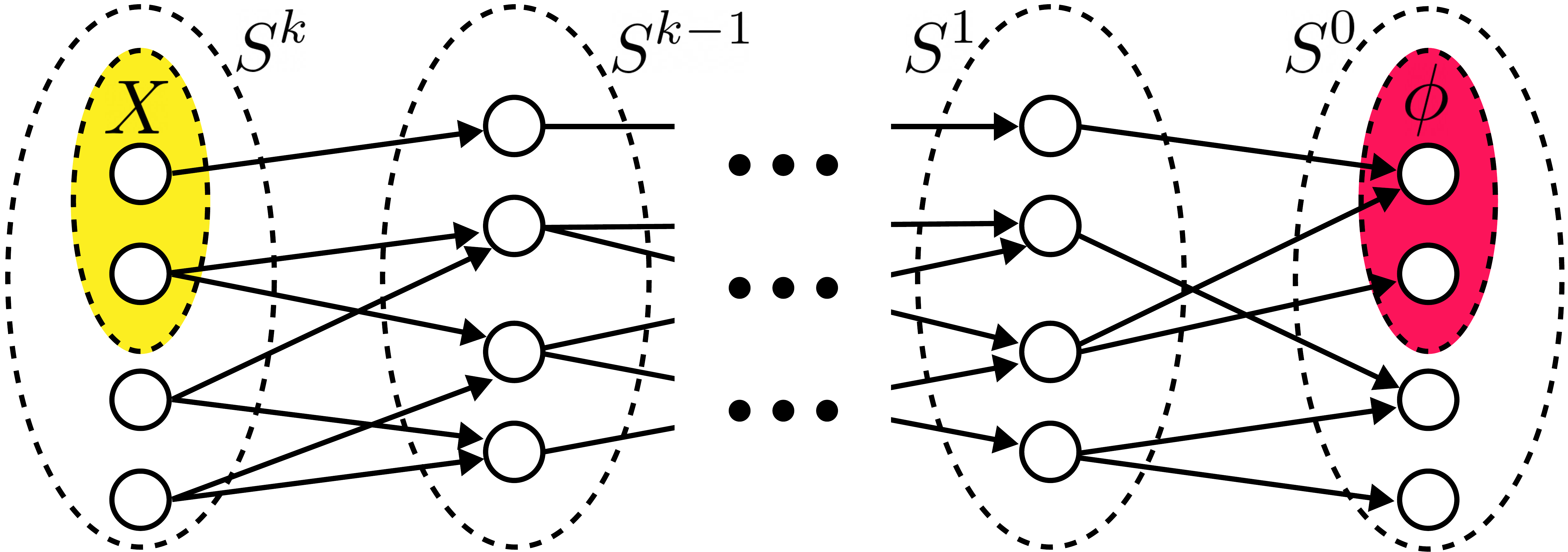}
    \vspace{-2mm}
    \caption{A layered causal graph.}
    \label{fig:lcg}
\end{figure}

A causal graph is a DAG where nodes are variables and edges denote cause-effect relations. A \emph{layered} causal graph (Fig.~\ref{fig:lcg}) is defined given an event $\phi$, for which we want to determine causes, and a set of variables $X \subseteq \mathcal{V}$, which we would like to evaluate as causal or not. A layered causal graph (LCG) is a DAG whose nodes are partitioned into non-intersecting layers $(S^k, \ldots, S^0)$, where for every edge $A \rightarrow B$ there exists some $i \in \{1, \ldots, k \}$ such that $A \in S^i$ and $B \in S^{i-1}$. Further, $X \subseteq S^k$, and $\phi \in S^0$.\looseness-1

\subsection{Markov Decision Processes}

A \emph{Markov decision process} (MDP) is a model for reasoning in fully observable, stochastic environments~\cite{bellman1966dynamic}, defined as a tuple $\langle S, A, T, R, d \rangle$. $S$ is a finite set of states, where $s \in S$ may be expressed in terms of a set of \emph{state factors}, $\langle f_1, f_2, \dots, f_N \rangle$, such that $s$ indexes a unique assignment of values to the factors $f$; $A$ is a finite set of actions; $T : S \times A \times S \rightarrow [0,1]$ represents the probability of reaching a state $s' \in S$ after performing an action $a \in A$ in a state $s \in S$; $R : S \times A \times S \rightarrow \mathbb{R}$ represents the expected immediate reward of reaching a state $s' \in S$ after performing an action $a \in A$ in a state $s \in S$; and $d : S \rightarrow [0, 1]$ represents the probability of starting in a state $s \in S$. A solution to an MDP is a policy $\pi : S \rightarrow A$ indicating that an action $\pi(s) \in A$ should be performed in a state $s \in S$. A policy $\pi$ induces a value function $V^\pi : S \rightarrow \mathbb{R}$ representing the expected discounted cumulative reward $V^\pi(s) \in \mathbb{R}$ for each state $s \in S$ given a discount factor $0 \leq \gamma < 1$. An optimal policy $\pi^*$ maximizes the expected discounted cumulative reward for every state $s \in S$ by satisfying the Bellman optimality equation $V^*(s) = \max_{a \in A} \sum_{s' \in S} T(s, a, s')[R(s, a, s') + \gamma V^*(s')]$.

\section{Structural Causal Models for MDPs}

At a high level, we construct a causal model of the computation that solves for the policy of an MDP and then use this model to determine causes for agent actions, which can later be used for explanation. In the general case, this process follows four steps. (1) A causal graph is generated from the relevant MDP components. (2) The resulting graph is converted into a layered causal graph. (3) The layered graph is pruned to remove any irrelevant nodes and edges, given $X$ and $\phi$. (4) A recursive algorithm identifies sets of causal variables in the pruned graph. This approach provides a principled, general framework for causal inference on MDPs while simultaneously supporting several types of explanations. 

In our layered graphical representation of MDPs we call layer $S^0$ the policy layer. This layer contains Boolean variables of the form $\pi_{sa} = [\pi(s) = a]$ that collectively represent the policy of the MDP. Thus, $S^0$ has $|S||A|$ variables, one for each state-action combination. Depending on the definition of $\mathcal{V}$, layers $S^1$-$S^k$ represent different parts of the MDP. In all definitions for $\mathcal{V}$ below we can derive the agent's action for some state $s$, given the layered causal graph for its MDP and the values of all nodes without incoming edges, by passing values along edges and computing variables in subsequent layers of the graph until we reach layer $S^0$. We will first detail this process in layered MDPs (\S4) and then discuss general MDPs and approximate methods (\S5).

\subsection{Causal Models for Layered MDPs}

We begin with the special case of layered MDPs, which contain both tree-structured MDPs and finite-horizon MDPs, and for which our methods are exact (up to discretization). Although it is possible to create a single, monolithic causal graph that simultaneously represents all components of the MDP tuple, this is not helpful since it does not afford any additional types of inference, is much less computationally efficient, and requires substantial bookkeeping to maintain the layered property. Thus, we analyze two causal models that, together, can answer causal queries about all parts of MDPs considered in previous literature. 

\begin{definition} 
A \emph{layered Markov decision process} is an MDP where, for all states $s \in S$, the successor graph of depth $h$ representing state $s$ is a layered graph $\forall h \in \mathbb{N}$.
\end{definition}

\paragraph{State Factors} The first model is for queries about the causality of state factors. Here, we let $\mathcal{U} = \emptyset$, and
\begin{equation*}
\mathcal{V} = \pi_{sa} \cup s \cup f_i \hspace{5mm} \forall s \in S, \forall a \in A, \forall i \in \{1, \ldots, n\}
\end{equation*}
\noindent where $f_i$ denotes the $i$th state factor. Finally, $\mathcal{M}$ is composed of the following three sets.
\begin{equation*}
    \mathcal{M}_F := f_i = f_i ^t, \hspace{5mm} \forall i \in \{1, \ldots, n\}.
\end{equation*}
\noindent Here $f_i ^t$ is the value of state factor $i$ at time $t$. A given set of state factors $\langle f_1, \ldots, f_n \rangle \in f$ determines the state $s \in S$.
\begin{equation*}
    \mathcal{M}_S := [s=s_i] = [f_1=f_1^i] \wedge \ldots \wedge [f_n=f_n^i], \hspace{5mm} \forall s \in S.
\end{equation*}

\noindent Last, we have equations representing action selection. 
\begin{equation*}
    \mathcal{M}_A := [\pi(s) = a] = \pi_{sa} \wedge s \hspace{5mm} \forall s \in S, a \in A.
\end{equation*}
\noindent Thus we define $~\displaystyle{\mathcal{M} := \mathcal{M}_F \cup \mathcal{M}_S \cup \mathcal{M}_A}$.

\begin{figure}[t]
    \centering
    \includegraphics[width=0.9\columnwidth]{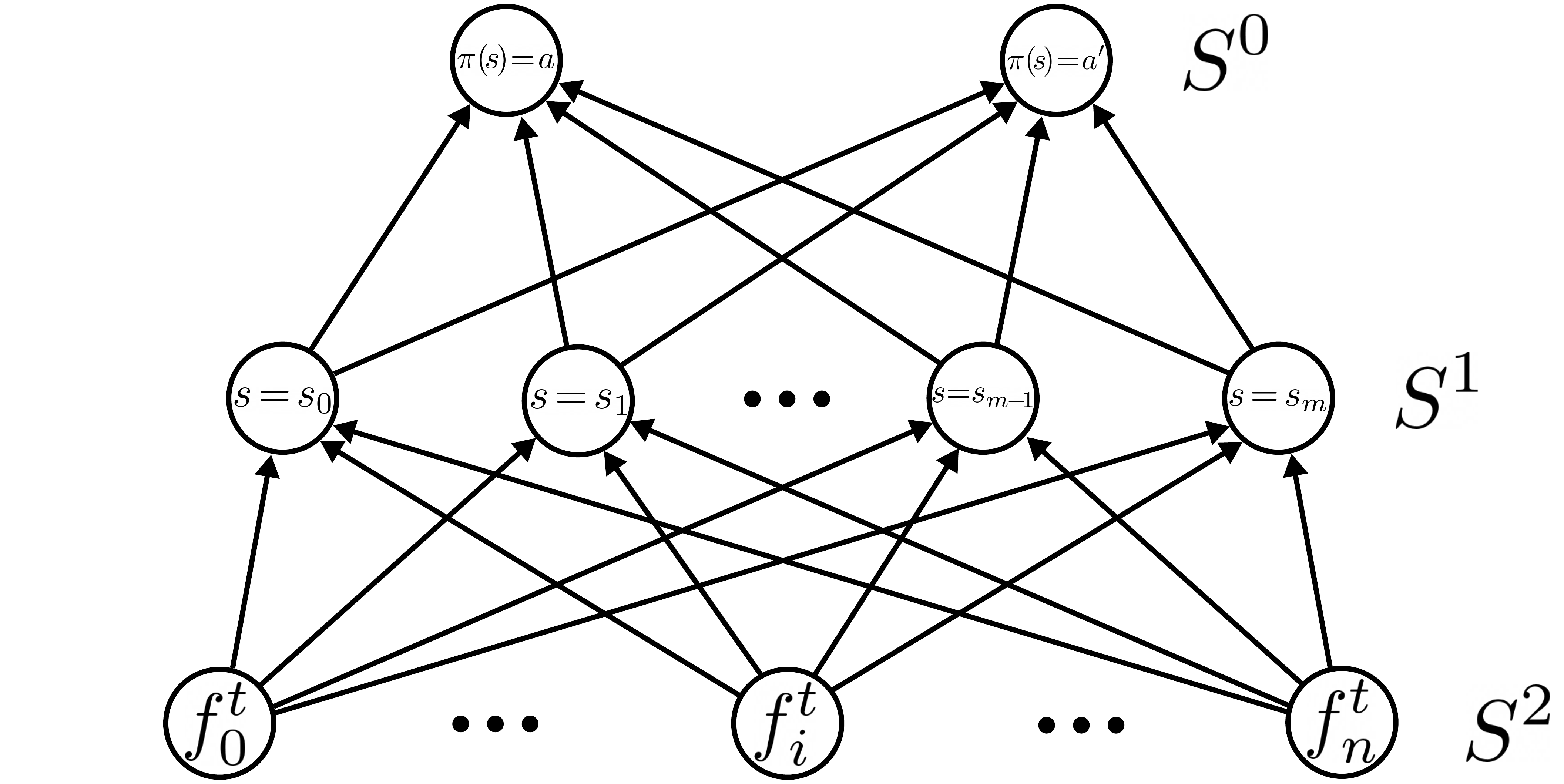}
    \caption{A layered causal graph generated from an MDP representing the influence of state factors on the action.}
    \vspace{-1mm}
    \label{fig:causaltreefactors}
    \vspace{4pt}
\end{figure}

Fig.\,\ref{fig:causaltreefactors} shows the causal graph represented by this SCM. In general, this definition of SCMs for state factors creates layered graphs with exactly three layers, and when the state space is discrete, permit exact inference regardless of the  underlying MDP topology. Importantly, this graph represents a \emph{fixed} policy. While we \emph{cannot} change state factors to produce a different policy, we \emph{can} understand how state factors affect action selection for a given policy.

\paragraph{Rewards, Transitions, and Values} The second causal model we present is used to analyze how reward, transition, and value functions causally influence action selection. Here, we let $\mathcal{U} = \{ \gamma \}$ since it is essential for computing the effect of other variables in the system, but we are unlikely to consider this a direct cause of any behavior we want to explain. Further, we let $\mathcal{V} = T(s, a, s') \cup R(s, a, s') \cup V(s) \cup \pi_{sa} ~~~ \forall s, a, s'$. Finally, we let $\mathcal{M}$ be the set of equations needed to solve for a policy, for instance by value iteration. The first two sets do not depend on other endogenous variables.
\begin{equation*}
    \mathcal{M}_R := R(s, a, s') = R^{ss'}_a, \hspace{5mm} \forall s,s' \in S, \forall a \in A;
\end{equation*}
\begin{equation*}
    \mathcal{M}_T := T(s, a, s') = T^{ss'}_a, \hspace{5mm} \forall s,s' \in S, \forall a \in A.
\end{equation*}

\noindent The set of equations for the value at each state $s \in S$ is
\begin{equation*}
    \mathcal{M}_V := V(s) = \max_{a} \sum_{s' \in S} T_a ^{ss'} [ R_a ^{ss'} + \gamma V(s') ], \hspace{5mm} \forall s \in S.
\end{equation*}

\noindent Last, we have the set of equations for action selection.
\begin{equation*}
    \mathcal{M}_A := (\pi(s) \hspace{-0.5mm} = \hspace{-0.8mm} a_k) \hspace{-0.5mm}=\hspace{-0.5mm} \Big[ \sum_{s' \in S} T_{a_k} ^{ss'} V(s') = A_{max} ^s \Big], \hspace{1mm} \forall s \in S.
\end{equation*}
\vspace{-2mm}

\noindent Here,
\vspace{-2mm}
\begin{equation*}
    A_{max} ^s = \max_{a} \Big ( \sum_{s' \in S} T_{a_1} ^{ss'} V(s'), \ldots, \sum_{s' \in S} T_{a_m}^{ss'} V(s') \Big ).
\end{equation*}
\noindent Thus we define $~\displaystyle{\mathcal{M} := \mathcal{M}_R \cup \mathcal{M}_T \cup \mathcal{M}_V \cup \mathcal{M}_A}$.

\begin{figure}[t]
    \centering
    \includegraphics[width=0.9\columnwidth]{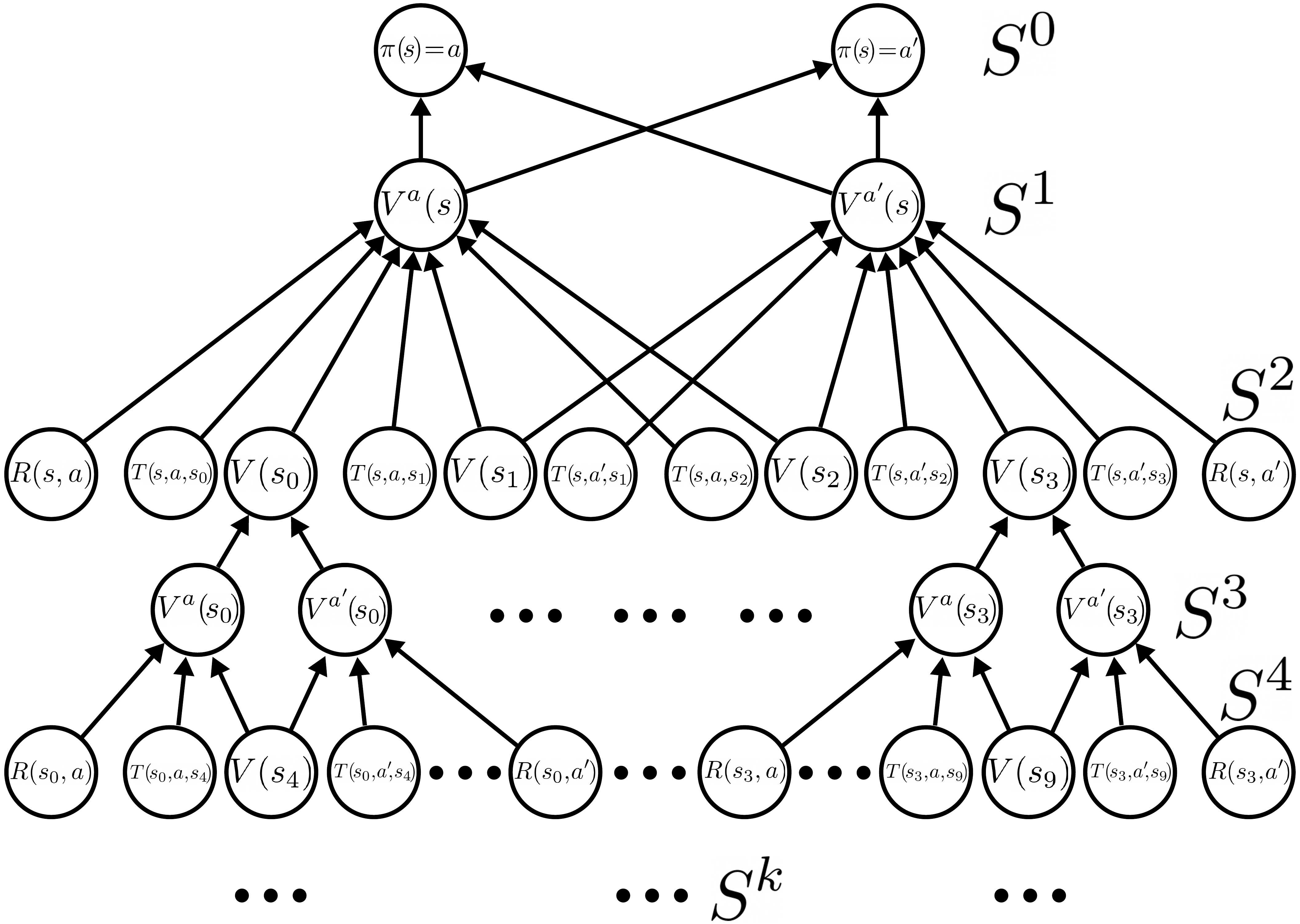}
    \vspace{-6pt}
    \caption{A layered causal graph representing the effect of rewards, transitions, and values on action selection. $V^a(s)$ is the value of taking action $a$ at state $s$.}
    \vspace{-1mm}
    \label{fig:causaltreemdp}
\end{figure}

The resultant LCG, shown in Fig.\,\ref{fig:causaltreemdp}, is built conditioned on the agent's current state. If the agent moves to a new state, a new graph is built, since reward and transition variables associated with successor states may change. Here, we also see that some layers contain value variables conditioned on particular actions ($V^a(s_i)$), allowing us to analyze arbitrary policies. If we want to answer queries about a fixed policy, these variables can be collapsed into the existing value variables ($V(s_i)$) since we no longer perform a max operation. Fig.\,\ref{fig:causaltreemdp} shows only 4 successor states in $S^2$, but there may be up to $|S|$. This graph thus reproduces either value iteration or policy evaluation depending on whether $V^a(s_i)$ variables are aggregated. It is also possible to move variables from $\mathcal{V}$ to the context $\mathcal{U}$, reducing complexity at the cost of eliminating variables from causal analysis. For example, we could move all reward variables to generate explanations using only transition variables.

\paragraph{Exactness Results} Layered MDPs are well-behaved since the SCMs formed by our construction naturally form LCGs. Given $\mathcal{M}$ and a state $s_0$, we can construct an LCG using the layered structure of the MDP. The following theorem and lemmas give a class of MDPs for which LCGs may be constructed and analyzed exactly.

\begin{theorem} 
Let $M$ be a finite-horizon MDP. If $G$ is a layered causal graph representing $M$ at state $s$, then $G$ preserves cause-effect relationships in the reasoning process for action selection in $M$ at $s$.
\end{theorem}

\begin{lemma}
Given a finite-horizon MDP with horizon $h$ and start state $s_0$, there exists an equivalent layered MDP.
\end{lemma}

\noindent \textbf{\textit{Proof of Lemma 1:}} We prove this by providing an algorithm for constructing layered MDPs from finite-horizon MDPs. For any start state $s_0$, we create a successor graph $H$ such that a directed edge from node $i$ to $j$ exists if and only if $\exists a \in A$ such that $T(s_i,a,s_j) > 0$.

Next, we run Breadth First Search without an explored list until all paths of length $\leq h$ have been explored and recorded. From these recorded paths, we can create a tree with root node $s_0$. We then append ``$k$" to state IDs at the $k$th level of the tree. Last, we aggregate any duplicate nodes, preserving their edges. Such nodes will only occur within the same layer of the tree. Thus, after aggregation, the resulting successor graph may not be a tree, but will be layered. \qed

\begin{lemma}
If $G$ and $H$ are causal graphs of finite Bayesian networks, and there exists a homomorphism $G \rightarrow H$, then $G$ and $H$ preserve cause-effect relationships.
\end{lemma}

\noindent \textbf{\textit{Proof of Lemma 2:}} This result follows from Jacobs et al. (\citeyear{jacobs2019causal}) and Otsuka and Saigo (\citeyear{otsuka2022equivalence}).

\noindent \textbf{\textit{Proof of Theorem 1:}} Since $M$ is finite-horizon, then by Lemma\,1 we can create an equivalent layered MDP and associated successor graph $H$ that represent action selection for any given state $s$. We can construct a function $\psi$ that induces a homomorphism $\psi : G \rightarrow H$ in the following way: map all nodes for variables $V(s_i)$ or $T(s, a, s_i)$ in $G$, to the node representing $s_i$ in $H$.\footnote{Fig.\,3 is an expanded version of the graph $G$, where the odd layers (representing max($\cdot$) operations) have been explicitly factored out to illustrate the possibility of modeling different policies.} Next, map all nodes for variables $R(s, a)$ to any node $n \in H$ such that $\psi(T(s, a, s_i)) = n$. Since the homomorphism $\psi: G \rightarrow H$ exists, and since MDPs may be represented as Bayesian networks, then by Lemma\,2, $G$ captures all cause-effect relationships for action selection. \qed

\subsection{Causal Inference for Layered MDPs}

Given an LCG, we can perform causal inference to determine causal variables with respect to an event $\phi$. In an MDP, a natural choice for $\phi$ is a subset of the variables $\pi_{sa}$. For example, if action $a$ is taken in state $s$ instead of $a'$, we have 
\begin{equation*}
    \phi = \langle [\pi(s) = a], [\pi(s) = a'] \rangle = \langle \textsc{True},\textsc{False} \rangle.
\end{equation*}
However, it is less clear how to define potential explanans, denoted by $X$. Intuitively, we often define $X$ as being, for example, the set of all state factors, the set of all reward variables, or the set of all values for states $h$ actions away. That is, we tend to define $X$ according to some semantic type. 

Given $X \subseteq \mathcal{V}$ and $\phi$, such that $\phi \cap X = \emptyset$, we would like to check if variables in $X$ cause $\phi$. We now review definitions of weak and actual cause from Halpern and Pearl (\citeyear{halpern2000causes}) and conditions for weak causality using LCGs. 

\begin{definition} 
Let $X \subseteq V$ be a subset of the endogenous variables, and let $x$ be a specific assignment of values for those variables. Given an event $\phi$, defined as a logical expression, for instance $\phi = (\neg a \wedge b)$, a \emph{weak cause} of $\phi$ satisfies the following conditions:

\begin{enumerate}
    \item Given the context $U=u$ and $X=x$, $\phi$ holds.
    \item Some $W \subseteq (V \setminus X)$ and some $\bar{x}$ and $w$ exist such that:
    
    \hspace{1mm} A) using these values produces $\neg \phi$.
    
    \hspace{1mm} B) for all $W' \subseteq W$, $Z \subseteq V \setminus (X \cup W)$, where $w'=w|W'$ and $z=Z$ given $U=u$, $\phi$ holds when $X=x$.
 
\end{enumerate}
\end{definition}

Essentially, item 2 b) is saying that, given context $U=u$, $X=x$ alone is sufficient to cause $\phi$, independent of some other endogenous variables $W$.

\begin{definition}
An \emph{actual cause} is a weak cause $X$ where no such $X' \subset X$ exists such that $X'$ is also a weak cause.
\end{definition}

We now introduce two additional constructs from Eiter and Lukasiewicz~(\citeyear{eiter2006causes}), that they use to establish a theorem on identification of weak causes. These constructs are general and do not have particular meaning with respect to MDPs. First, however, some notation. $\mathcal{P}(\cdot)$ and $\mathcal{D}(\cdot)$ denote power set and domain, respectively. $\phi_{xy}(u)$ is the value of $\phi$ given context $U=u$ and the assignments of variables $X=x$ and $Y=y$. $w' = w|W'$ refers to the restriction of $w$ to $W'$. Last, $[x \langle y] = x|(X \setminus Y ) \cup y$. 
\begin{align*} 
    R^0= & \{ (\mathbf{p}, \mathbf{q}, F) | F \subseteq S^0, \mathbf{p}, \mathbf{q} \subseteq \mathcal{D}(F), \hspace{75pt} \\
    & \exists w \in \mathcal{D}(S^0 \setminus F) \forall p, q \in \mathcal{D}(F): \\
    & p \in \mathbf{p} \text{ iff } \neg \phi_{pw}(u), \\
    & q \in \mathbf{q} \text{ iff } \phi_{[q \langle \hat{Z}(u)]w'}(u) \\
    & \forall W' \subseteq S^0 \setminus F, w'=w|W', \hat{Z} \subseteq F \setminus S^k, 
\end{align*}
\ \hspace{19pt} and
\begin{align*}
    R^i= & \{ (\mathbf{p}, \mathbf{q}, F) | F \subseteq S^i, \mathbf{p}, \mathbf{q} \subseteq \mathcal{D}(F), \\
    & \exists w \in \mathcal{D}(S^0 \setminus F) \exists (\mathbf{p'}, \mathbf{q'}, F') \in R^{i-1} \forall p, q \in \mathcal{D}(F): \\
    & p \in \mathbf{p} \text{ iff } F_{pw}'(u) \in \mathbf{p'}, \\
    & q \in \mathbf{q} \text{ iff } F_{[q \langle \hat{Z}(u)]w'}'(u) \in \mathbf{q'} \\
    & \forall W' \subseteq S^0 \setminus F, w'=w|W', \hat{Z} \subseteq F \setminus S^k, \text{ for } i > 0.
\end{align*}

\begin{algorithm}[t] \small
  \begin{algorithmic}[1]
  \STATE \textbf{Input:} Layered causal graph $G$, variables $X$, event $\phi$
  \STATE \textbf{Output:} Sets of weak causal variables $\mathcal{C}_W \subseteq X$ of $\phi$.
  \STATE $R^0 \gets \emptyset$, $S^0, S^k \gets$ layers of $G$ containing $\phi, X$
  \FORALL{$F \in \mathcal{P}(S^0)$}
  \STATE $\bf{p}, \bf{q} \gets \emptyset$
  \FORALL{$w \in \mathcal{D}(S^0 \setminus F)$}
  \FORALL{$p \in \mathcal{D}(F)$}
  \IF{$\neg \phi$ given $p$ and $w$}
  \STATE ${\bf p} \gets p \cup {\bf p}$
  \ENDIF
  \ENDFOR
  \FORALL{$q \in \mathcal{D}(F)$}
  \STATE $b \gets$ \textsc{True}
  \FORALL{$W' \in \mathcal{P}(S^0 \setminus F)$}
  \STATE $w' \gets w|W'$
  \FORALL{$\hat{Z} \in \mathcal{P}(F \setminus S^k)$}
  \STATE $z' \gets \hat{Z}(u)$
  \IF{$\neg \phi$ given $q$, $\hat{z}$, and $w'$}
  \STATE $b \gets$ \textsc{False}
  \STATE {\bf break}
  \ENDIF
  \ENDFOR
  \IF{$\neg b$}
  \STATE {\bf break}
  \ENDIF
  \ENDFOR
  \IF{$b$}
  \STATE ${\bf q} \gets q \cup {\bf q}$
  \ENDIF
  \ENDFOR
  \ENDFOR
  \STATE $R^0 \gets ({\bf p}, {\bf q}, F) \cup R^0$
  \ENDFOR
  \STATE $R^- \gets R^0$
  \STATE $l \gets 1$
  \WHILE{$l \leq k$}
  \STATE $R \gets$ \textsc{RecurrenceStep}$(S^l, S^k, R^-)$
  \STATE $R^- \gets R$, $l \gets l+1$
  \ENDWHILE
  \STATE $\mathcal{C}_W \gets \emptyset$
  \FORALL{$({\bf p}, {\bf q}, F) \in R^-$}
  \IF{${\bf p} \not = \emptyset$ \textbf{and} $x \in {\bf q}$}
  \STATE $\mathcal{C}_W \gets x \cup \mathcal{C}_W$
  \ENDIF
  \ENDFOR
  \RETURN $\mathcal{C}_W$
  \end{algorithmic}
  \caption{\textsc{Determine Weak Causes}}
  \label{alg:idwc}
\end{algorithm}
\begin{theorem}
 (From Eiter and Lukasiewicz (\citeyear{eiter2006causes})) Let $\mathcal{S} = (\mathcal{U}, \mathcal{V}, \mathcal{M})$ be a causal model. Let $X \subseteq \mathcal{V}$, $x \in \mathcal{D}(X)$, $u \in \mathcal{D}(U)$, and let $\phi$ be an event. Let $(S^0, \ldots, S^k)$ be a layering of $G(\mathcal{S})$ relative to $X$ and $\phi$, and let $R^k$ be defined as above. Then, $X=x$ is a weak cause of $\phi$ under $u$ in $\mathcal{S}$ iff 
 ~(1)~$X(u)=x$ and $\phi(u)$ in $\mathcal{S}$, and
 ~(2)~$\exists (p,q,X) \in R^k$ such that $\mathbf{p} \not = \emptyset$ and $x \in \mathbf{q}$.
\end{theorem}
We now develop a naive, exact algorithm, Algorithm \ref{alg:idwc}, that computes $R^0, \ldots, R^k$ and then applies Theorem 2 to determine weak causality. At a high level, Algorithm \ref{alg:idwc} proceeds up the causal chain in the LCG, recursively identifying causes one layer at a time. That is, sets of variables in $S^1$ are identified as causes of events in $S^0$. Those variables then assume the role of the event(s) and their causes are identified in $S^2$. This process repeats until the algorithm reaches layer $S^k$. Lines 7-9 check condition 2A from Def. 2, while lines 10-22 check condition 2B. Condition 1 is always satisfied since $\phi$ represents some part of the agent's actual policy. The recurrence step (pseudocode in Appendix A) applies the same reasoning to the output of the initial step. The final result is a family of sets of causal variables $\mathcal{C}_W$, where $\mathcal{C}_W ^i \subseteq X$, that each satisfy Def. 2 w.r.t. the original event $\phi$. We direct interested readers to Eiter and Lukasiewicz (\citeyear{eiter2006causes}) for a thorough treatment of Theorem 2 and definitions of $R^0$ and $R^k$. 

One challenge is the continuous domain of variables representing the transition and reward functions. Here, we assume a discretization scheme. For example, reward variables could have discrete domains bounded by the min and max of the original reward function. Furthermore, the variables in $X$ must be located in the same layer in the causal graph.

Thus, Algorithm \ref{alg:idwc} supports analysis of LCGs as in Figs.\,\ref{fig:causaltreefactors} and \,\ref{fig:causaltreemdp}, but is not efficient and does not exploit any structure in MDPs. If we restrict $X$ to the state factors in the MDP, we get LCGs as in Fig.\,\ref{fig:causaltreefactors}. Here, we can use a simpler algorithm, similar to that presented by Bertossi et al.~(\citeyear{bertossi2020causalitybased}), based on the concepts of responsibility and blame from Chockler and Halpern (\citeyear{chockler2004responsibility}). Algorithm \ref{alg:countweakcause} iterates directly through possible weak causal sets (line 4), and then progressively checks larger sets $W$ for assignments $w$ that satisfy Def. 2. Lines 11-15 and 19-22 check conditions 2A and 2B, respectively. Finally, lines 23-27 compute the responsibility score, $\rho$, used to determine if $F$ is weakly causal. In addition to finding weak causal sets consistent with Def. 2, $\rho$ provides a ranking over causal sets. Algorithm \ref{alg:countweakcause}, \textsc{MeanRESP}, has some properties that allow us to expand its applicability later.

\begin{algorithm}[t] \small
  \begin{algorithmic}[1]
  \STATE \textbf{Input:} State factors $F$, states $S$, policy $\pi$, start state $s_0$
  \STATE \textbf{Output:} Set of weak causes $\mathcal{C}_W$, responsibility scores $\mathcal{R}$.
  \STATE $\mathcal{C}_W, \mathcal{R} \gets \emptyset$
  \FORALL{$F' \in \mathcal{P}(F)$}
  \FORALL{$\beta = 0 ... |F \setminus F'|$}
  \STATE $\sigma, t \gets 0$
  \FORALL{$W \in \mathcal{P}(F \setminus F')$ such that $|W| = \beta$}
  \FORALL{$w \in \mathcal{D}(W)$}
  \STATE $b \gets \textsc{True}$
  \STATE $t \gets t+1$
  \FORALL{$W' \in \mathcal{P}(W)$}
  \STATE $w' \gets w|W'$
  \STATE $s' \gets [s_0 \langle w' ]$
  \IF{$\pi(s') \neq \pi(s_0)$}
  \STATE $b \gets \textsc{False}$
  \STATE \textbf{break}
  \ENDIF
  \ENDFOR
  \IF{$\neg b$}
  \STATE \textbf{continue}
  \ENDIF
  \FORALL{$f' \in \mathcal{D}(F')$ such that $s_0|F' \not = f'$}
  \STATE $s' \gets [s_0 \langle (f' \cup w)]$
  \IF{$\pi(s') \neq \pi(s_0)$}
  \STATE $\sigma \gets \sigma + \frac{1}{|\mathcal{D}(F')|}$
  \ENDIF
  \ENDFOR
  \ENDFOR
  \ENDFOR
  \STATE $\rho \gets \sigma/(t(1+\beta))$
  \IF{$\rho > 0$}
  \STATE $\mathcal{C}_W \gets \mathcal{C}_W \cup F'$
  \STATE $\mathcal{R}$.\textsc{Append}$(\rho)$
  \STATE \textbf{break}
  \ENDIF
  \ENDFOR
  \ENDFOR
  \RETURN $\mathcal{C}_W$
  \end{algorithmic}
  \caption{\textsc{MeanRESP}}
  \label{alg:countweakcause}
\end{algorithm}

We can determine actual causes by checking the minimality condition within the weak causes $\mathcal{C}_W$ (psuedocode in Appendix A). Further, since causal queries are made with respect to an event $\phi$ and variables $X$, inference may be sped up by removing irrelevant variables. Graphs absent such variables are called \emph{strongly reduced} (see the removal conditions and algorithm in Appendix A).

\section{Generalization and Approximation}

Although layered MDPs encompass a large class of MDPs, Algorithms \ref{alg:idwc} and \ref{alg:countweakcause} have three key limitations. (1) They cannot represent infinite horizon problems. (2) While the graph itself is straightforward to build for finite horizon problems, very large problems or problems with large horizons may still be prohibitively expensive to analyze. (3) These algorithms cannot handle continuous state factors. In this section we propose several modifications to \textsc{MeanRESP} and an additional pre-processing step that address these limitations, either by constructing smaller, approximate causal models or by approximating more expensive inference processes.

\subsection{Approximate Causal Models for MDPs}

Here, we address limitation (1). There are many methods for building approximate causal graphs of arbitrary MDPs, depending on the available information. We assume the original graph is built using a generic, uninformed algorithm, such as the one we present in Appendix A based on Iwasaki et al. (\citeyear{iwasaki1986causality}). The resulting causal graph may not be unique if $\mathcal{M}$ contains circular dependencies \cite{iwasaki1986causality,trave1997causal}. Since the Bellman update equation 
is a recurrence relation between MDP state values, non-layered structures are highly likely.

Thus, we develop Algorithm \ref{alg:decycle} which, given state $s_0$ and causal graph $G$, removes these structures to produce an LCG $G_{s_0}$ for causal analysis whenever the agent is in state $s_0$. We consider a horizon $h$ and let variables associated with states not reachable within $h$ actions form causal `leaves' by removing their incoming causal edges. Remaining non-layered structures are corrected by removing edges such that states farther from $s_0$ causally influence states nearer to $s_0$, forming a simplified, finite-horizon version of the original MDP. These operations are executed simultaneously in Algorithm \ref{alg:decycle}. In a sense, the causal influence flows from the horizon at time $t + h$ back in time to the present time $t$. 

\begin{algorithm}[t!] \small
  \begin{algorithmic}[1]
  \STATE \textbf{Input:} Causal graph $G$, successor graph $H$, current state $s_0$, horizon $h$
  \STATE \textbf{Output:} Layered causal graph $G_{s_0}$
  \STATE $G_{s_0}(E',\mathcal{V}) \gets G(E,\mathcal{V})$
  \FORALL{$V_{s_i} \subset \mathcal{V}$ where $V_{s_i}$ represents variables for state $s_i$}
  \IF{$s_i$ is reachable on $H$ in $\kappa \leq h$ actions from $s_0$}
  \STATE label all $v \in V_i$ with $\kappa$
  \ELSE
  \STATE label all $v \in V_i$ with $\infty$
  \ENDIF
  \ENDFOR
  \FORALL{$v \in \mathcal{V}$}
  \IF{label$(v) = \infty$}
  \STATE remove all incoming edges to $v$
  \STATE \textbf{continue}
  \ENDIF
  \FORALL{$v' \in \mathcal{V}$}
  \IF{label$(v) \geq$ label$(v')$} 
  \STATE $E' \gets E' \setminus$ \{ directed edges from $v'$ to $v$ \}
  \ENDIF
  \ENDFOR
  \ENDFOR
  \RETURN $G_{s_0}$
  \end{algorithmic}
  \caption{\textsc{Construct Layered Causal Graph}}
  \label{alg:decycle}
\end{algorithm}

\subsection{Approximate Causal Inference for MDPs}

Often, models are too large for exact inference. Moreover, we may wish to apply more restrictive versions of Def. 2, or extend analysis to real-valued state factors. We address these problems via tweaks to \textsc{MeanRESP}. If state factors are real-valued or their domains are simply too large, we can approximate inference by replacing lines 7 and 8 with a single for loop over vectors $w$ of size $\beta$ generated via sampling. Input-output pairs are constructed and counted in the same way and the responsibility score still indicates weak causality. The challenge becomes determining sampling domains that cover important counterfactual scenarios efficiently.

There are several techniques applicable to both discrete and continuous domains. One class involves limiting the sizes of $W$ and $Z$, the benefit of which is reduction in problem complexity at the cost of omitting potential weak causes. These restrictions of course diverge from Def. 2, but can be made in a principled way that more or less preserves an order over the possible results. In particular, one may set $Z=\emptyset$ and or $|W| \leq \eta$ for some $\eta < < |\mathcal{V}|$. In the state factor case, the latter restriction is equivalent to requiring $\rho \geq \epsilon > 0$ rather than $\rho > 0$, for weak causality. Moreover, while we use one Def. 2, Halpern and Pearl, and later Halpern~\cite{halpern2015modification}, have proposed several definitions.

Often, reward or transition variables in MDPs are not independent, but instead depend on a high-level rule. For example, reward may be proportional to the value of a state factor, or the transition function may encode identical slipping probabilities regardless of location, as in classic grid-world domains. These rules constrain the transition and reward function to manifolds, and we can discretize these manifold to gain efficiency without sacrificing important possible worlds. Given such structure we can replace loops over, for example, all possible values of $T(s, a, s')$, and instead loop over the set of high-level rules, which is much smaller. 

Finally, we may want to perform causal analysis on value function variables of future states. If we want to look far into the future, or the branching factor of the successor graph is large, the resulting LCG may be too large to analyze, even when limiting domains or $|W|$ and $|Z|$. In these cases, we can use a form of beam search to limit the intermediate events represented at each layer of the LCG. The idea is to measure the influence of variables on $\phi$ and then keep only the $m$ most influential variables as the search progresses.

\section{Results}
We present three results: 1) a case study highlighting our framework's flexibility, 2) results of a user study showing strong user preference for explanations generated via causal analysis, 3) asymptotic run time and memory use bounds, and a discussion on the tightness of these bounds in practice and potential pre-computation.

\subsection{Case Study: Explanation Diversity}

The purpose of this study is to show how (1) our approach can handle \emph{semantically different} types of causal queries, corresponding to different conceptions of MDP explanation in the literature, and (2) formal definitions of causality identify sensible explanans. There are many templates for inserting variables identified as explanans into natural language, which is the predominant method. Here, we first qualitatively examine the correctness of causal attribution.

\setlength{\tabcolsep}{4.5px}
\begin{table}[t!] \scriptsize
\centering
\begin{tabular}{|c|c|c|c|c|c|} 
\hline
Method & $F$ & $R$ & $T$ & $V$ & Causal? \\
\hline\hline
\citeit{elizalde2009generating} & Yes & - & - & - & No \\
\hline
\citeit{russell2019explaining} & Yes & - & - & - & No \\
\hline
Khan et al. (\citeyear{khan2009minimal}) & - & Yes & - & - & No \\
\hline
\citeit{juozapaitis2019explainable} & - & Yes & - & - & No \\
\hline
Betram et al. (\citeyear{bertram2018explainable}) & - & Yes & - & - & No \\
\hline
Wang et al. (\citeyear{wang2016impact}) & - & - & Yes & - & No \\ 
\hline
\citeit{madumal2020explainable} & Yes & Yes & - & - & Yes \\
\hline
\textbf{Proposed} & \textbf{Yes} & \textbf{Yes} & \textbf{Yes} & \textbf{Yes} & \textbf{Yes} \\
\hline
\end{tabular}
\captionof{table}{Comparison of method applicability} 
\vspace{-5mm}
\label{tab:methods}
\end{table}

We have identified 4 general types of explanation in the literature, each focusing on one component of the MDP tuple: \emph{state factors} ($F$)~(Elizalde et al. \citeyear{elizalde2009generating}; Russell et al. \citeyear{russell2019explaining}), \emph{rewards} ($R$)~(Khan et al. \citeyear{khan2009minimal}; Juozapaitis et al. \citeyear{juozapaitis2019explainable}; Bertram and Wei \citeyear{bertram2018explainable}), \emph{transitions} ($T$)~(Wang et al. \citeyear{wang2016impact}), and \emph{future states and values} ($V$)~\cite{pouget2020ranking}.  These papers define metrics, algorithms, and definitions particular to their type, and lead us to define the following.

\begin{definition}
$Y$\!-type explanations use explanans $x \subset Y$. For example, $F$\!-type explanations use the set of state factors.
\end{definition}

\begin{figure}[t]
    \centering
    \includegraphics[width=0.70\columnwidth]{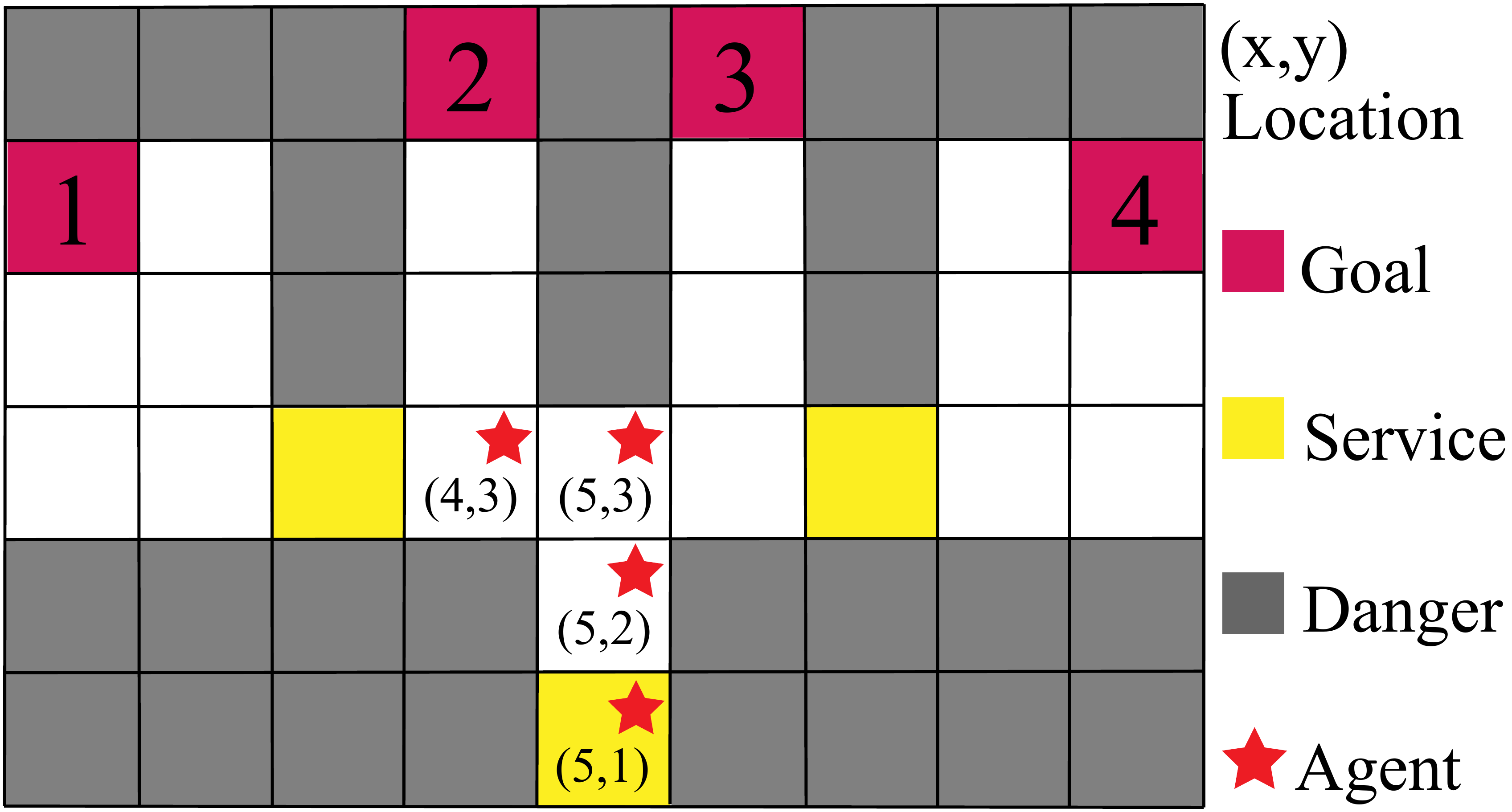}
    \vspace{-3mm}
    \caption{Example domain. The agent begins at $(5,1)$.}
    \vspace{-2mm}
    \label{fig:grid_domain}
\end{figure}

We now detail an example MDP domain. Consider a robot navigating the environment depicted in Figure\,\ref{fig:grid_domain}. The agent knows: its $(x, y)$ location, time to failure, $c$, and if its location is normal, ideal for repairs, or hazardous, $t$. Thus, the state factors are $x \in \{1, \ldots, 9\}$, $y \hspace{-0.7mm} \in \hspace{-1mm} \{1, \ldots, 6\}$, $c \hspace{-0.7mm} \in \hspace{-1mm} \{0, \ldots, 5\}$, $t \hspace{-0.7mm} \in \hspace{-1mm} \{ \hspace{-0.4mm} \textsc{Normal(N)}, \textsc{Repair(R)}, \textsc{Danger(D)} \hspace{-0.4mm} \}$. The actions are $A \hspace{-0.7mm} = \hspace{-0.7mm} \{ \textsc{Up}, \textsc{Left}, \textsc{Right}, \textsc{Repair} \}$. If the agent breaks down and cannot be repaired or visits a hazardous state, it gets a reward of $-10$. Repairing has a reward of $R_C$, and reaching the $i$th goal state yields reward $R_i$. All other state-actions have a reward of $-1$. Last, transitions are deterministic except for actions \textsc{Left} and \textsc{Right} when taken at $(5, 3)$, which result in agent locations of $(4, 3)$ (with probability $T_L \hspace{-1mm} = \hspace{-1mm} 0.6$) or $(4, 4)$, and $(6, 3)$ (with probability $T_R \hspace{-1mm} = \hspace{-1mm} 0.01$) or $(6, 4)$, respectively. For all examples we considered event $\phi \hspace{-1mm} = \hspace{-1mm} \pi_{s_0}$, and for explanation type $Y$ we set $X\hspace{-1mm}=\hspace{-1mm}Y$ and apply \textsc{MeanRESP}. Table \ref{tab:scenarios} shows the value of all state factors $s_0$, rewards, and actions for each scenario. Below, we briefly contextualize \textsc{MeanRESP}'s output.

\textbf{Scenario 1:} The cause for action \textsc{Repair} is $t=$ R ($\rho = 0.64$). Since $R_C>0$, it is optimal to repair as it prevents failure later and is better than the default reward of $-1$.

\textbf{Scenarios 2 and 3:} There are two causes of action \textsc{Up} in scenario 2: $t\hspace{-0.5mm}=$N ($\rho \hspace{-0.5mm}= \hspace{-0.5mm}0.50$) and $(x,y)\hspace{-0.5mm}=$(5,2) ($\rho\hspace{-0.5mm} =\hspace{-0.5mm} 0.26$). Scenario 3 has the same causes, but $\pi(s_0)\hspace{-0.8mm}=$\textsc{Left} and $(x,y)$ has $\rho\hspace{-0.5mm} = \hspace{-0.5mm}0.60$. This is due to the topology at (5,3) compared to (5,2). In both cases, $t$ is a cause since if $t\hspace{-0.5mm}=$\textsc{R}, $\pi(s_0)\hspace{-0.5mm}=$\textsc{Repair}.

\textbf{Scenario 4:} The causes for action \textsc{Left} are $c\hspace{-0.5mm}=\hspace{-0.5mm}2$ ($\rho = 0.85$), $(x,y)\hspace{-0.5mm}=$(4,3) ($\rho = 0.68$), and $t\hspace{-0.5mm}=$\textsc{N} ($\rho = 0.60$). Note that unlike in previous scenarios, time to failure is both a cause and has the highest responsibility score. If the agent were to go directly to goal 2, it will break down at (4,5).

\textbf{Scenario 5:} There are two causes of action \textsc{Left}: $R_1$ ($\rho \hspace{-0.5mm} = \hspace{-0.5mm} 0.30$) and $R_3$ ($\rho \hspace{-0.5mm} = \hspace{-0.5mm} 0.55$). Since we bound $\mathcal{R} = [-100, 100]$, no values for $R_2$ or $R_4$ can change the outcome. $R_4$ is already the maximum, and $R_2$ alone is not a cause due to its relatively weak effect on the expected value of that subtree and transition variables were in $\mathcal{U}$ and not $\mathcal{V}$.

\textbf{Scenario 6:} Here, only $R_3$ ($\rho \hspace{-1.1mm} = \hspace{-1.1mm} 0.55$) is causal. Since $R_C\hspace{-1mm} > \hspace{-1mm}0$, the agent exploits this by repeatedly taking service at (3,3). Thus, $R_1$ alone cannot affect the agent's policy since goals 1 and 4 will never be visited. This is a good example of the proposed approach identifying a possible poorly specified objective.

\textbf{Scenarios 7 and 8:} Since $R_1\hspace{-1mm}=\hspace{-1mm}R_2$ in scenario 7, the only cause of the action \textsc{Left} is $T_R$ ($\rho\hspace{-0.5mm}=\hspace{-0.5mm}0.66$). However, in scenario 8, both $T_R$ ($\rho\hspace{-0.5mm}=\hspace{-0.5mm}0.66$) and $T_L$ ($\rho\hspace{-0.5mm}=\hspace{-0.5mm}0.56$) are causes.

\textbf{Scenario 9:} Using \textsc{MeanRESP} with beam search, we find that the most influential set of trajectories lead to goal 1. That is, its value contributes most to the expected value, even though the most likely ($p{=}0.6$) outcome of taking action \textsc{Left} at (5,3) reaches goal 2.

\setlength{\tabcolsep}{4.5px}
\begin{table}[t!] \scriptsize
\centering
\begin{tabular}{|c|c|c|c|c|c|c|c|c|} 
\hline
ID & $Y$ & $s_0$ $(x,y,t,c)$ & $R_1$ & $R_2$ & $R_3$ & $R_4$ & $R_C$ & $Action$ \\
\hline\hline
1 & $F$ & (5,1), R, 5  & 80 & -50 & -40 & 100 & 2 & \textsc{Repair} \\ 
\hline
2 & $F$ & (5,2), N, 4 & 80 & -50 & -40 & 100 & 2 & \textsc{Up} \\
\hline
3 & $F$ & (5,3), N, 3 & 80 & -50 & -40 & 100 & 2 & \textsc{Left} \\
\hline
4 & $F$ & (4,3), N, 2 & 80 & 90 & -40 & 100 & -1 & \textsc{Left} \\
\hline
5 & $R$ & (5,3), N, 3 & 80 & -50 & -40 & 100 & -1 & \textsc{Left} \\
\hline
6 & $R$ & (5,3), N, 3 & 80 & -50 & -40 & 100 & 2 & \textsc{Left} \\
 \hline
7 & $T$ & (5,3), N, 3 & 80 & 80 & 70 & 100 & -1 & \textsc{Left} \\
\hline
8 & $T$ & (5,3), N, 3 & 80 & -50 & -40 & 100 & -1 & \textsc{Left} \\
\hline
9 & $V$ & (5,3), N, 3 & 80 & -50 & -40 & 100 & -1 & \textsc{Left} \\
\hline
\end{tabular}
\captionof{table}{List of scenarios} 
\vspace{-4mm}
\label{tab:scenarios}
\end{table}

\textbf{Summary:} These scenarios show how different sets of explanans provide \emph{semantically distinct} insights into variables' effects on actions. This underscores the utility of flexibility in generating explanations since the `best' explanans may be unknown pre-deployment. Because existing methods only analyze one component of the MDP, they cannot produce most of these explanations (Table \ref{tab:methods}).

\subsection{User Study}

There is no definitive, automated metric for the quality of an explanation. Therefore, the gold standard for assessing the effectiveness of an explanation is a user study. Here, we describe the results of such a study, investigating the following three hypotheses. 

\begin{itemize}
    \item \textbf{H1:} Users tend to prefer explanations generated using causal reasoning over explanations generated using heuristics.
    \item \textbf{H2:} Users tend to prefer explanations supported by explanans representing specific types of information.
    \item \textbf{H3:} User preferences for explanation methods or explanan types correlate with demographic or lifestyle indicators.
\end{itemize}

\subsubsection{Study Description and Administration}

In total, $189$ participants from the United States and Canada were recruited via the crowd-sourcing platform Prolific (www.prolific.co). Participants were fluent in English, were aged between 18 and 65, and 49\% identified as male, 46\% as female, and 5\% as non-binary.

\begin{figure}[t]
    \centering
    \includegraphics[width=0.9\columnwidth]{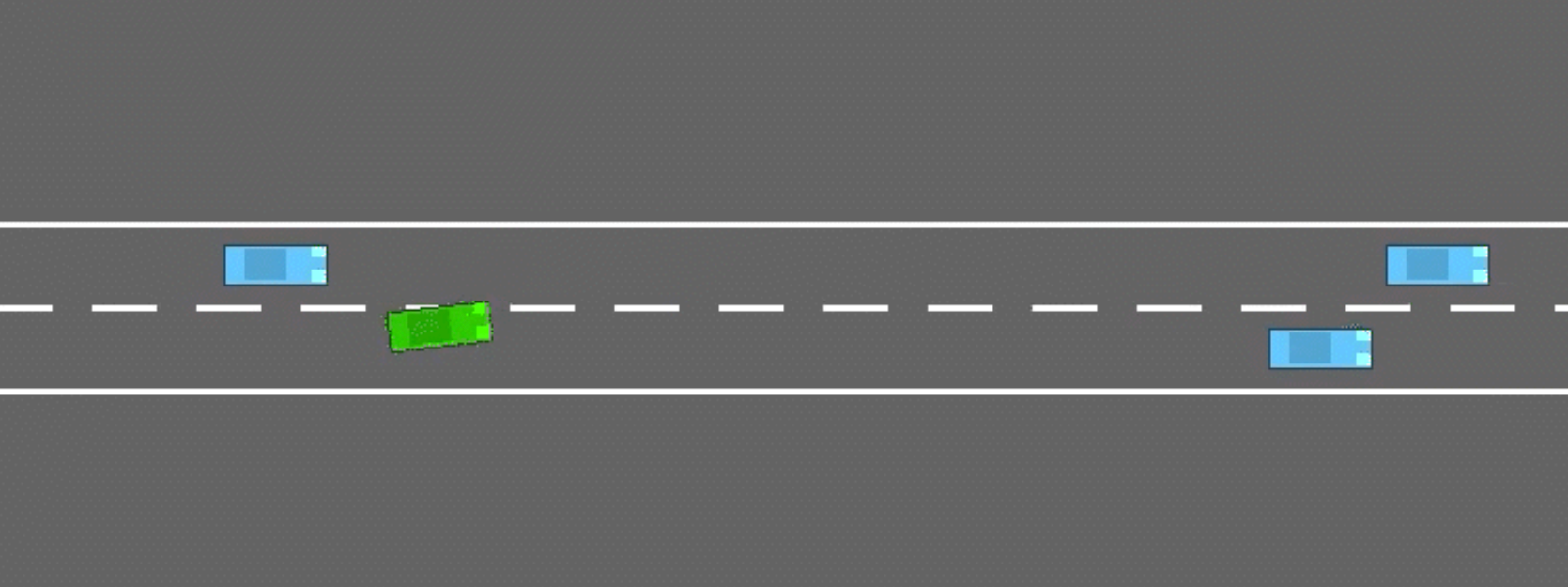}
    \vspace{-6pt}
    \caption{An example scene from the simulator where the ego car (green) makes a left lane change.}
    \label{fig:simenv}
\end{figure}

Participants were shown three short clips of simulated driving scenarios in a randomized order, where a car drives on a highway \cite{highway-env} and changes speeds and lanes based on a policy from an MDP. After each particular action, such as the left lane change shown in Fig.~\ref{fig:simenv}, participants were shown 7 different explanations in a randomized order, each generated by a different method for automatic explanation, and asked to rank them relative to each other to produce a strict preference ordering. The explanations shown included three baselines: \cite{elizalde2009generating} ($F$-type), \cite{khan2009minimal} ($R$-type), and \cite{wang2016impact} $T$-type as well as all four types of explanation generated by our proposed method.

To present as little bias towards different explanations as possible, every explanation was presented using the same basic template: "The car <took action> because <explanan 1>, ..., <explanan N>." Each explanan in the MDP had a custom phrase, signifying both what the explanan represented and its value. 

\subsubsection{Study Results}

We find remarkably strong evidence in support of \textbf{H1} and \textbf{H2}, and surprisingly, strong evidence in favor of the null hypothesis w.r.t. \textbf{H3}. Below we review the results in detail.

\paragraph{Preferences for Causal Explanations}

\begin{figure}[h]
    \centering
    \vspace{-2pt}
    \includegraphics[width=0.9\columnwidth]{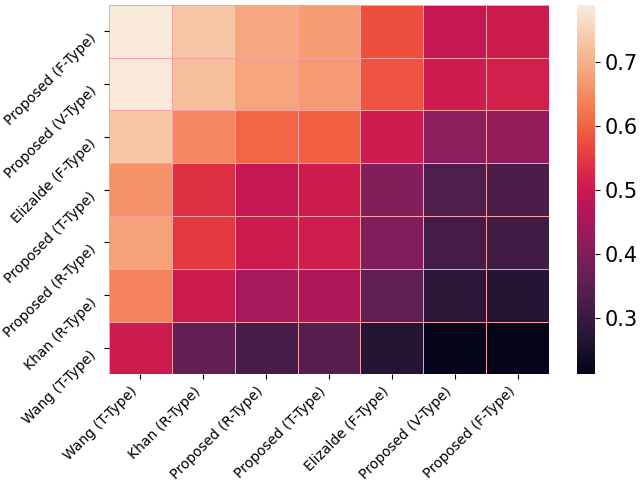}
    \vspace{-8pt}
    \caption{Preference likelihoods for MDP explanation methods. The color of cell $(row, col)$ indicates the probability that explanations generated using method $row$ are preferred to explanations generated from method $col$.}
    \label{fig:r1}
\end{figure}

Figure \ref{fig:r1} summarizes our findings on user preferences for explanation methods. The most important observation is that, for every explanation type, users prefer the explanations generated via causal reasoning over those generated via heuristic methods. We applied the Mann-Whitney U-test~\cite{mann1947test} to each pair of generation methods (21 in total), using an initial $\alpha$-value of $0.5$, and a Bonferonni corrected~\cite{bonferroni1936teoria} $\alpha$-value of $0.0024$. We detected the following preference ordering with p-values below 0.0001.

\begin{center}
1) Prop-$F$ $\sim$ Prop-$V$ $\succ$ Elizalde $\succ$ Prop-$R$ $\sim$ Prop-$T$ $\succ$ Khan $\succ$ Wang

\end{center}

\noindent Here, $A \succ B$ denotes a strict preference for $A$ over $B$, and $\sim$ denotes preference equality.

We believe the overall preference for causal explanations is due to their consistent relevance across all scenarios. For example, although both heuristic and causal $F$-type methods have access to the same potential explanans, and occasionally produce the same explanations, there are some cases where the heuristic methods do not produce sensible explanations, such as the following:

\noindent \textbf{Heuristic Example:} The car changed lanes to the right because the car was in the left lane.

\noindent \textbf{Casual Reasoning Example:} The car changed lanes to the right because the car was in the left lane, the estimated time to collision in the left lane was 2 seconds, and the right lane was empty.

Clearly, the heuristic method fails to provide both relevant and complete information to explain the event in this case.

\paragraph{Preferences for Explanan Types}

\begin{figure}[t]
    \centering
    \vspace{-12pt}
    \includegraphics[width=0.48\columnwidth]{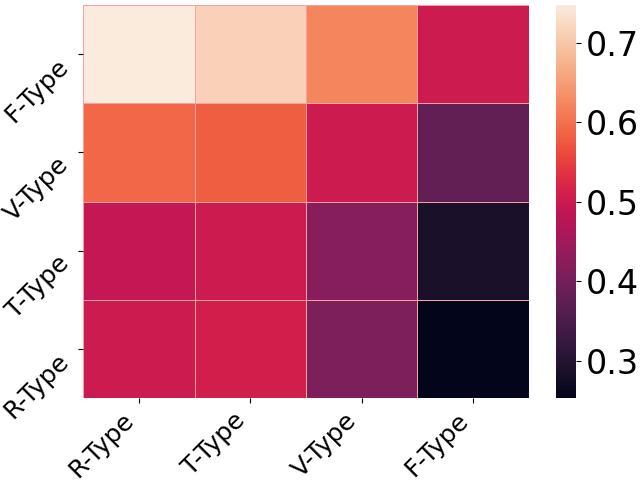}
    \vspace{-12pt}
    \caption{Preference likelihoods for MDP explanations based on explanan type.}
    \label{fig:r3}
\end{figure}

Figure \ref{fig:r3} shows a similar analysis for the different types of explanation ($F$,$R$,$T$,$V$), where we can see that $F$-type explanations are, on average, preferred over $R$-type and $T$-type. For this analysis if there were multiple explanations based on the same explanans we used the most preferred rank. For example, if a user preferred $A \succ B \succ C$, where $A$ and $C$ were $F$-type and $B$ was $R$-type, then we would record $F$-type as being ranked highest, followed by $R$-type. We apply the same pair-wise Mann-Whitney analysis as in {\bf H1}, now with a total of 6 pairs and a Bonferonni corrected $\alpha$-value of 0.0083. Following this analysis, we obtain the following preference ordering: $F$-type $\succ$ $V$-type $\succ$ $R$-type $\sim$ $T$-type with p-value 0.00001.

\paragraph{Demographic and Lifestyle Non-Impact}
We found these results to be consistent across genders, age groups, and rates of technology use. We also, somewhat surprisingly, found these results to be consistent regardless of the frequency with which participants operated motor vehicles. We also tried several unsupervised clustering methods to check for more complex correlations between user demographics and preferences for different types of explanations, but did not find anything significant. This should give practitioners confidence that these results will hold in many settings.

\subsection{Resource Use and Practical Application}

Table \ref{tab:complexity} shows some preliminary bounds on resource use. Here, $|S|$ is the size of the state space, $|F|$ is the number of state factors, and $|\mathcal{V}|$ is the number of endogenous variables.

In practice, worst-case bounds are loose. In Algorithm~\ref{alg:idwc}, we check all assignments of $X$, and $X \subseteq \mathcal{V}$. However, usually $|X| << |\mathcal{V}|$, especially after reducing the LCG. Bounds for Algorithm \ref{alg:countweakcause} are also poor estimates of in-practice cost, since it uses short-circuiting. Some bounds' tightness depends on the connectivity of the MDP. For Algorithm \ref{alg:decycle}, the bounds assume fully-connected MDPs, but most MDPs are sparse and thus the number of edges $E << |S|^2$. Moreover, if $h$ is small compared to the width of the MDP, run time will decrease since nodes labeled $\infty$ are handled in linear time.

There are other possible improvements since theoretically every explanation can be pre-computed, but this is impractical due to the number of possible explanations. Notably, constructing LCGs for each state, regardless of how $\phi$ and $X$ are specified, and computing connectivity and reachability allows reductions and causal model approximations to be applied quickly online, given $X$ and $\phi$. 

\vspace{-1mm}
\setlength{\tabcolsep}{4.5px}
\begin{table}[h] \scriptsize
\centering
\begin{tabular}{|r|l|c|l|} 
\hline
Alg. & Res. & Complexity & Bottleneck \\
\hline\hline
1 & time & $2^{|\mathcal{V}|}|\mathcal{D}(\mathcal{V})|^2 k$ & Enumerating causal sets \\ 
 & space & $4|\mathcal{V}|2^{|\mathcal{V}|}$ & Storing $R$, $R^-$ \\
\hline
2 & time & $2^{2|F|}|\mathcal{D}(F)|$ & Enumerating causal sets \\ 
 & space & $2^{|F|}|F|$ & Storing weak causes \\
\hline
3 & time & $|S|^2 (h + 1)$ & Connectivity checks +~Label comparisons \\
 & space & $|S|^2$ & Reachability matrix \\
\hline
\end{tabular}
\captionof{table}{Worst-case resource bounds for algorithms 1-3.}
\vspace{-8mm}
\label{tab:complexity}
\end{table}

\section{Conclusion}

We present a novel framework for causal analysis of MDPs using SCMs, motivated by generating explanations of MDP agent behavior. Principally, this framework provides (1) a theoretical foundation for explainable sequential decision making, and (2) simultaneous support for causal queries using different decision problem components, which has previously not been possible. Future work will empirically investigate user preferences for explanations generated using causal models compared to heuristics and how different contexts affect user preferences for types of explanation.

\section*{Appendix A}

\subsection{Producing Causal Graphs}

Algorithm \ref{alg:construct} generates a causal graph given a structural causal model by first constructing a bipartite graph, where variables ($\mathcal{V}$) and equations ($\mathcal{M}$) are nodes, and edges exist between variable nodes and equation nodes if that equation contains that variable. Given the bipartite graph, Hopcroft-Karp is run to produce a perfect matching. This perfect matching is used to build a \emph{directed} (causal) graph containing only variables.

\begin{algorithm}[h] \small
  \begin{algorithmic}[1]
  \STATE \textbf{Input:} Set of variables $\mathcal{V}$, set of equations $\mathcal{M}$
  \STATE \textbf{Output:} Causal graph $G$
  \STATE $\mathcal{B} \gets$ \textsc{ConstructBipartite}$(\mathcal{V}, \mathcal{M})$
  \STATE $E_{PM} \gets$ \textsc{Hopcroft-Karp}$(\mathcal{B})$
  \STATE $V \gets \mathcal{V}$, $E \gets \emptyset$
  \FORALL{$v \in \mathcal{V}$}
  \STATE \text{//} $Q$ is a node in $\mathcal{B}$ representing an equation.
  \FORALL{$e(v, Q) \in \textsc{Edges}(v)$} 
  \IF{$e \in E_{PM}$}
  \STATE \text{//} $V_Q$ is the set of variables in equation $Q$.
  \FORALL{$v' \in V_Q, v' \not = v$} 
  \STATE $E \gets E \cup \textsc{Edge}(v', v)$
  \ENDFOR
  \ELSE
  \FORALL{$v' \in V_Q, v' \not = v$}
  \STATE $E \gets E \cup \textsc{Edge}(v, v')$
  \ENDFOR
  \ENDIF
  \ENDFOR
  \ENDFOR
  \STATE $G \gets \{E, V \}$
  \RETURN $G$
  \end{algorithmic}
  \caption{\textsc{Construct Causal Graph}}
  \label{alg:construct}
\end{algorithm}

\subsection{Reducing Causal Graphs}

Eiter and Lukasiewicz~(\citeyear{eiter2006causes}) provide the following conditions for removing a variable $v$ from a causal graph.

\begin{enumerate}
    \item $v \in X$ is not connected via variables in $V \setminus X$ to $\phi$.
    \item $v$ is neither a direct parent of a variable in $\phi$ nor part of a chain connecting $X$ to $\phi$.
\end{enumerate}

Given an LCG $G_{s_0}$ Algorithm~\ref{alg:prune} produces a strongly reduced LCG $G_{s_0}^{\phi X}$.

\begin{algorithm}[h] \small
  \begin{algorithmic}[1]
  \STATE \textbf{Input:} Layered causal graph $G_{s_0}$, explanans $X$, event $\phi$
  \STATE \textbf{Output:} Strongly reduced layered causal graph $G_{s_0}^{\phi X}$
  \STATE $G_{s_0}^{\phi X} \gets G_{s_0}$
  \FORALL{$x \in X$}
  \IF{$\not \exists$ path from $x$ to some $y \in \phi$}
  \STATE remove $x$ and its edges from $G_{s_0}^{\phi X}$
  \ENDIF
  \IF{$\forall$ paths from $x$ to $\phi$, $\exists x' \in X$ along the path}
  \STATE remove $x$ and its edges from $G_{s_0}^{\phi X}$
  \ENDIF
  \ENDFOR
  \FORALL{$v \in V \setminus (X \cup \phi)$}
  \IF{$\not \exists y \in \phi$ such that $v$ is a direct parent \textbf{and} $\not \exists x \in X, y \in \phi$ such that $v \in x \rightarrow y$}
  \STATE remove $v$ and its edges from $G_{s_0}^{\phi X}$
  \ENDIF
  \ENDFOR
  \RETURN $G_{s_0}^{\phi X}$
  \end{algorithmic}
  \caption{\textsc{Reduce Causal Graph}}
  \label{alg:prune}
\end{algorithm}

\subsection{Recurrence Step for \textsc{Determine Weak Causes}}

The outer loop (line 4) looks at all possible subsets of variables, $F$, in the $i$th layer. Variables not in $F$ are assigned values $w$ one at a time, eventually looping over all possible sets of values (line 6). Then, for every tuple $R^-$ from layer $i-1$ (line 7), we check the conditions for $p \in \bf{p}$ (lines 8-10) and $q \in \bf{q}$ (lines 11-23). Finally, for a given set $F$, we add all the qualifying $p, q$ to the tuple $R$ (line 24).

\begin{algorithm}[h] \small
  \begin{algorithmic}[1]
  \STATE \textbf{Input:} Layers $S^i$, $S^k$, tuples $R^-$
  \STATE \textbf{Output:} Set of tuples $R$.
  \STATE $R \gets \emptyset$
  \FORALL{$F \in \mathcal{P}(S^i)$}
  \STATE $\bf{p} \gets \emptyset$; $\bf{q} \gets \emptyset$
  \FORALL{$w \in \mathcal{D}(S^i \setminus F)$}
  \FORALL{$({\bf p'}, {\bf q'}, F') \in R^-$}
  \FORALL{$p \in \mathcal{D}(F)$}
  \IF{$F'$, given $p$ and $w$, is in ${\bf p'}$}
  \STATE ${\bf p} \gets p \cup {\bf p}$
  \ENDIF
  \ENDFOR
  \FORALL{$q \in \mathcal{D}(F)$}
  \STATE $b \gets$ \textsc{True}
  \FORALL{$W' \in \mathcal{P}(S^i \setminus F)$}
  \STATE $w' \gets w|W'$
  \FORALL{$\hat{Z} \in \mathcal{P}(F \setminus S^k)$}
  \STATE $z' \gets \hat{Z}(u)$
  \IF{$F'$, given $q$, $z'$, and $w'$, is {\bf not} in {\bf $q'$}}
  \STATE $b \gets$ \textsc{False}
  \STATE {\bf break}
  \ENDIF
  \ENDFOR
  \IF{$\neg b$}
  \STATE {\bf break}
  \ENDIF
  \ENDFOR
  \IF{$b$}
  \STATE ${\bf q} \gets q \cup {\bf q}$
  \ENDIF
  \ENDFOR
  \ENDFOR
  \ENDFOR
  \STATE $R \gets ({\bf p}, {\bf q}, F) \cup R$
  \ENDFOR
  \RETURN $R$
  \end{algorithmic}
  \caption{\textsc{Recurrence Step}}
  \label{alg:rdwc}
\end{algorithm}

\bibliographystyle{ACM-Reference-Format}
\balance
\bibliography{bibliography}

\end{document}